\documentclass[review]{elsarticle}

\journal{Journal of Pattern Recognition}

\usepackage{lineno,hyperref}

\usepackage{graphicx}
\usepackage{amsmath}
\usepackage{amsfonts}
\usepackage{algorithm, algorithmic}
\usepackage{subfigure}
\usepackage{booktabs}
\usepackage{geometry}
\usepackage{multirow}
\selectfont

\begin{document}

\begin{frontmatter}

\title{A Graph Data Augmentation Strategy with Entropy Preservation}

\author[ad1,ad2]{Xue Liu}
\author[ad3,ad4]{Dan Sun}
\author[ad3,ad1,ad4,ad5]{Wei Wei\corref{cor1}}\ead{weiw@buaa.edu.cn}

\address[ad1]{Institute of Artificial Intelligence, Beihang University, Beijing, 100191, China}
\address[ad2]{Beijing System Design Institute of Electro-Mechanic Engineering, Beijing, 100854, China}
\address[ad3]{School of Mathematical Sciences, Beihang University, Beijing, 100191, China}
\address[ad4]{Key Laboratory of Mathematics, Informatics and Behavioral Semantics, Ministry of Education, 100191, China}
\address[ad5]{Peng Cheng Laboratory, Shenzhen, Guangdong, 518066, China}

\cortext[cor1]{Corresponding author}

\begin{abstract}
The Graph Convolutional Networks (GCN) proposed by Kipf and Welling is an effective model for semi-supervised learning, but faces the obstacle of over-smoothing, which will weaken the representation ability of GCN. Recently some works are proposed to tackle above limitation by randomly perturbing graph topology or feature matrix to generate data augmentations as input for training. However, these operations inevitably do damage to the integrity of information structures and have to sacrifice the smoothness of feature manifold. In this paper, we first introduce a novel graph entropy definition as a measure to quantitatively evaluate the smoothness of a data manifold and then point out that this graph entropy is controlled by triangle motif-based information structures. Considering the preservation of graph entropy, we propose an effective strategy to generate randomly perturbed training data but maintain both graph topology and graph entropy. Extensive experiments have been conducted on real-world datasets and the results verify the effectiveness of our proposed method in improving semi-supervised node classification accuracy compared with a surge of baselines. Beyond that, our proposed approach could significantly enhance the robustness of training process for GCN.
\end{abstract}

\end{frontmatter}

\section*{Introduction}
Graph, as a ubiquitous data structure, is employed extensively in a wide range of applications, such as cheminformatics~\cite{molecular_graph}, interactive mechanism analysis~\cite{interactive_mechanism} and social networks~\cite{social_networks}. All of these domains and many more can be readily modeled as graphs, which contain information about the connection between individual units. For instance, the citation graph, as an academic interactive network, describes interactions among science research papers which are represented as nodes with labels to indicate category, and the citation links between papers are mapped into edges. Information from a single node or local dense nodes propagates along edges, and this makes graphs useful structured knowledge repositories for machine learning tasks like link prediction and node classification.

Graph Convolutional Networks (GCN)~\cite{GCN} draws support from convolutional operation on a graph to aggregate neighbor nodes information from low- to high-order hierarchical structures to get central node representation. The feed-forward propagation in GCN model consists of $k$ layers of graph convolution, which is similar to perception but additionally has a neighborhood aggregation step motivated by spectral convolution.

In order to enable GCN with more expressivity to wider neighbors, one may stack more layers on the network. But unfortunately, the deeper layer network model fails to achieve the expectation partly due to the phenomenon of over-smoothing~\cite{over-smoothing_2}, which is an inherent issue of graph convolutional calculation mechanism. It has been proven that graph convolution operation is a type of Laplacian smoothing, i.e., the higher power operation of normalized adjacency matrix, thus representations of nodes in the same region converge to same value and tend to be indistinguishable across different classes in embedding space as the model goes deeper~\cite{DeeperInsight-of-GCN}.

An easy but effective way to tackle over-smoothing is to generate perturbed data for training by randomly deleting elements from the adjacent matrix or feature matrix. But this graph perturbation strategy inevitably breaks the integrity of substructures that are vital to graph topology or information propagation. And more precisely, these perturbation procedures inevitably result in the damage to the smoothness of the data manifold, which is also the theoretical basis of semi-supervised learning tasks on GCN. Thus how to quantify such smoothness and how to preserve such smoothness when generating augmentations still need attention.

As a fundamental concept of statistical physics and information theory, graph entropy is commonly used to quantitatively measure the dynamics~\cite{EntropyDynamic_2} and describe the change of graph topology as well as graph features. In this paper, we propose a new graph entropy as an index to describe the smoothness of the graph feature manifold and yield that the key point to control this kind of smoothness lies in the triangle motif-based information structures. Afterward, a novel graph data perturbation strategy for the over-smoothing problem of GCN is provided, whereby graph entropy could be preserved as much as possible. The main steps in this augmentation strategy are as follows. Firstly, we tend to keep the original adjacent matrix unchanged instead of dropping any nodes or edges from the input graph for each training epoch. Then nodes from specific shape motifs and nodes not in motifs but selected with a certain probability are set as activated status. Only activated nodes' features could be present in the feature matrix while the remaining are reset as zero vectors. In this study, we focus on the triangle motifs for their ubiquitousness in understanding the interaction of social networks and their contribution to the preservation of graph entropy. Extensive experiments have been conducted on several real-world datasets and the results demonstrate the effectiveness of our proposed method in reducing over-smoothing and increasing robustness during the whole training process. In addition, our results significantly improve semi-supervised nodes classification performance compared to state-of-the-art methods. We summarize the main contributions as follows.
\begin{itemize}
\item[(1)] We provide a new graph entropy to measure the smoothness of the graph feature manifold and conclude that the motif-based information structures determine this graph entropy.
\item[(2)] We propose a novel graph data augmentation strategy that protects not only the integrity of topological structure but also the integrity of motif-based information units. Our strategy shows an advantage in the preservation of graph entropy compared with other methods.
\item[(3)] Extensive experiments are conducted on several real-world datasets to show the effectiveness of our proposed method.
\item[(4)] Our approach significantly enhances the robustness of GCN and could alleviate the over-smoothing phenomenon to a certain extent.
\end{itemize}

The rest part of this paper is organized as the following. The basic concepts and related works are introduced in Section~\ref{section-2}. The newly defined graph entropy is provided in Section~\ref{section-3}. The brief introduction of methodology is presented in Section~\ref{section-4}. The theoretical analysis of our method could be seen in Section~\ref{section-5}. The results of our experiments are provided in Section~\ref{section-6}. Our conclusions are summarized in Section~\ref{section-7}.

\section{Background}\label{section-2}
Let $G = (\mathcal{V}, \mathcal{E})$ denote a graph with node set $\mathcal{V}$ and edge set $\mathcal{E} \subseteq \mathcal{V}\times \mathcal{V}$, $G$ has a feature matrix $\mathbf{X} \in \mathbb{R}^{|\mathcal{V}| \times \eta}$ with $i$-th row $\mathbf{X}_{i}$ corresponding to the feature vector of node $v_{i}$ with length $\eta$, and training labels for all nodes are listed in $\mathbf{Y} \in\{0,1\}^{|\mathcal{V}| \times c}$, where $c$ is the classes number and each row $\mathbf{Y}_{i}$ of $\mathbf{Y}$ denotes the label of node $v_{i}$. The adjacency matrix $\mathbf{A} \in \mathbb{R}^{|\mathcal{V}| \times |\mathcal{V}|}$ encodes the node-wise connection of the network, whose entry $\mathbf{A}_{ij} = 1$ if there exits an edge between node $v_i$ and $v_j$, otherwise $\mathbf{A}_{ij} = 0$.

\subsection{Graph Convolutional Networks}
Graph Convolutional Networks (GCN) generalizes neural techniques into graph-structured data. The core operation in GCN is graph propagation, in which information spreads from each node to its neighbors with some deterministic propagation rules. The feed-forward propagation in the GCN model consists of $k$ layers of graph convolution, which is recursively conducted as
\begin{equation}\label{GCN}
     \quad \mathbf{H}^{(l)}=\left\{
     \begin{aligned}
      \sigma(\hat{\mathbf{A}}\mathbf{H}^{(l-1)}\mathbf{W}^{(l-1)}) &,\ if\ l\in[1,...,k] \\
      \mathbf{X}&,\ if\ l=0
      \end{aligned}
      \right..
\end{equation}
Here $\hat{\mathbf{A}}=\hat{\mathbf{D}}^{-1/2}(\mathbf{A}+\mathbf{I})\hat{\mathbf{D}}^{-1/2}$ is a symmetrically normalized adjacency matrix with self-connections, where $\hat{\mathbf{D}}$ is the degree matrix of $\mathbf{A}+\mathbf{I}$ and $\mathbf{I}$ denotes the identity matrix. $\mathbf{H}^{(l)} = \{h_1^{(l)},...,h_{|V|}^{(l)}\}$ represents the hidden vectors of the $l$-th layer with $h^{(l)}_{i}$ as the hidden features of node $v_{i}$. $\sigma(\cdot)$ denotes a nonlinear function, and $\mathbf{W}^{(l)}$ is the corresponding weight matrix for the $l$-th layer.

\subsection{Related Works in Alleviating Over-smoothing}
\begin{figure*}[htbp]
\centering
\includegraphics[height=7cm,width=15cm]{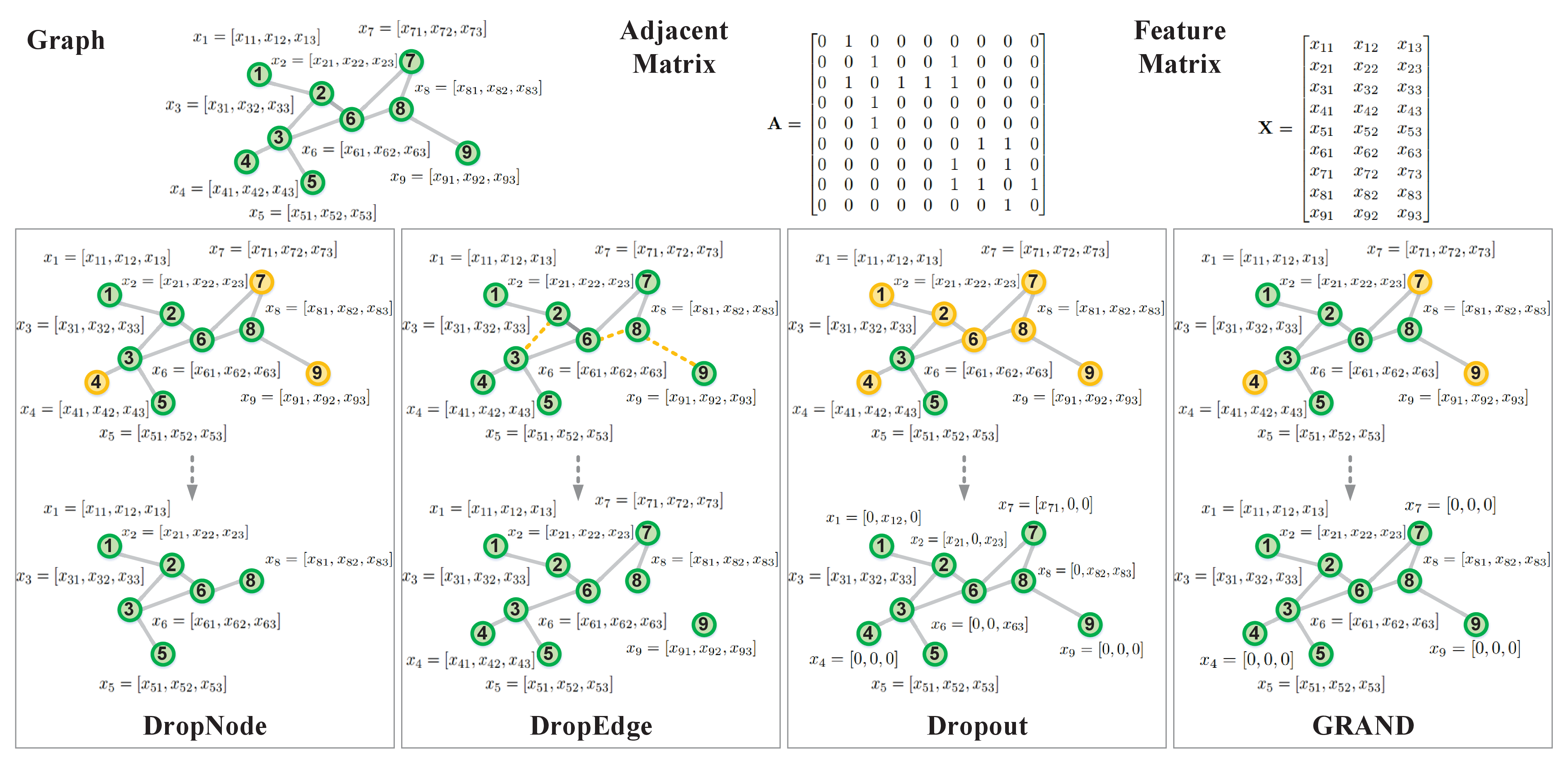}
\caption{
The illustration of four graph data augmentation approaches (i.e., DropNode, DropEdge, Dropout, and GRAND) that help alleviate over-smoothing for GCN.
}
\label{Figure-InstructureForGraphAugMethods}
\end{figure*}

Recently, a series of related works are proposed to alleviate the over-smoothing phenomenon of GCN, and most of them take the approach of perturbing graph data for training. DropNode~\cite{AS-GCN}, DropEdge~\cite{Dropedge}, Dropout~\cite{Dropout}, and GRAND~\cite{GRAND} are four typical tricks, which are shown in Figure~\ref{Figure-InstructureForGraphAugMethods}.

DropNode and DropEdge belong to the topology-based perturbation approaches, while Dropout and GRAND are in the category of graph feature-based perturbation methods. In detail, DropNode samples subgraphs for mini-batch training by randomly removing a part of nodes according to proportion $p$ as well as edges connected to the dropped nodes. As a consequence, this method will construct a subgraph $SG$ of the original graph $G$, satisfying $\mathcal{V}(SG)\subseteq \mathcal{V}(G), \mathcal{E}(SG)\subseteq \mathcal{E}(G)$. DropEdge acts as a data augmenter by randomly dropping a certain rate of edges from the input graph. Formally, it randomly enforces $|\mathcal{E}_{p}|$ non-zero elements of the adjacent matrix $\mathbf{A}$ to be zeros, where $\mathcal{E}_{p}$ is the dropped edges set selected by probability $p$. Dropout tries to perturb the feature matrix by randomly setting some elements in feature matrix $\mathbf{X}$ to be zeros, i.e., $\tilde{\mathbf{X}}_{ij} = \frac{\epsilon_{ij}}{1-\delta}\mathbf{X}_{ij}$, where $\mathbf{X}_{ij}$ is the $j$-th element of the $i$-th row vector $\mathbf{X}_{i}$ in feature matrix $\mathbf{X}$, and $\epsilon_{ij}$ draws from Bernoulli distribution $\mathbf{B}(1-\delta)$ parameterized by droprate $\delta$. GRAND randomly sets some nodes' features to be zero vectors, i.e., $\tilde{\mathbf{X}}_{i} = \frac{\epsilon_{i}}{1-\delta}\mathbf{X}_{i}$, where $\mathbf{X}_{i}$ denotes the $i$-th row vector of feature matrix $\mathbf{X}$ and $\epsilon_{i}$ draws from Bernoulli distribution $\mathbf{B}(1-\delta)$.

However, these methods inevitably break the smoothness of the data manifold in the perturbation procedures of randomly deleting elements from topology or features. It is worth mentioning that the smooth manifold constitutes the fundamental of semi-supervised learning tasks for GCN.

\section{Graph Entropy}\label{section-3}
In this part, we use the concept of entropy to measure the smoothness of the graph data manifold. Entropy is a fundamental law of statistical physics, and the second law of thermodynamics shows that the entropy of a macroscopic system is hard to decrease. Shannon introduced the concept of entropy into information theory as a characteristic measure to reveal information related to a system. As representations of complex systems, real networks are usually very large, and one can characterize graph information quantitatively in terms of macroscopic parameters using methods similar to entropy. Thus graph entropy is widely used to describe and understand the dynamics of a graph quantitatively in terms of general topology or features. It was first introduced by Rashevsky~\cite{GraphEntropyBeginging}, then Mowshowitz investigated graph entropy to measure the structural information content of graphs~\cite{entropyAndComplexity} and Körner applied a different definition of graph entropy into coding theory~\cite{korner1973coding}.

Most graph entropies are derived from the basic Shannon's entropy definition, whose details are as follows. For a discrete system $\mathcal{X}$, $I(x_{i})=-\log p(x_{i})$ denotes the self-information of $x_{i}\in \mathcal{X}$ with occurring probability $p(x_{i})$. The entropy of system $\mathcal{X}$ is defined by $H(\mathcal{X})$, as
\begin{equation}\label{Shannon_Entropy}
     H(\mathcal{X})= - \sum\limits_{i=1}^{|\mathcal{X}|} p(x_{i})\log p(x_{i}).
\end{equation}

Usually, information-theoretic measures for graphs are based on a graph invariant and then derive a partitioning~\cite{entropyAndComplexity}. Instead of determining partitions of elements based on a given invariant, Dehmer et al. developed an approach that was based on using so-called information functional $f$, mapping sets of nodes to the positive reals~\cite{entropy_and_function}, via
\begin{equation}\label{information_functional}
     p(x_{i}) = \frac{f(x_{i})}{\sum\limits_{j=1}^{|\mathcal{X}|}f(x_{j})}.
\end{equation}
Then graph entropy measure is obtained by applying functions (\ref{Shannon_Entropy}) and (\ref{information_functional}).

Graph entropy measures the randomness or uncertainty from a statistical perspective. Maximum entropy description retains all of the uncertainty not removed from the original data, and it has been interpreted as the maximally noncommittal concerning missing information~\cite{MaxEntropy_2}. Here we briefly present a novel graph entropy design in terms of features on nodes as well as neighborhoods relations to evaluate the diffusion of global feature information.

\subsection{Smoothness Index}\label{section_Smoothness_Index}

In this part, we provide a new graph entropy to indicate the smoothness of the global information distribution. Its idea comes from an application of entropy in image segmentation, in which each pixel of a digital image maps to nodes and one divides them into different communities based on image contrasts. Entropy plays a significant role in quantifying the smoothness of the texture in various regions of image analysis: high entropy indicates more smoothness of the texture and less abrupt graphic blocks. As a consequence, more information will be contained in the target image since it exhibits a more uniform distribution~\cite{image_entropy}.

In our new graph entropy design, feature vector of each node is regarded as an individual, and then all of them constitute a feature vector space. In particular, in accordance with previous definition, $\mathbf{X} \in \mathbb{R}^{|\mathcal{V}| \times \eta}$ denotes the feature matrix for graph $G=(\mathcal{V}, \mathcal{E})$, where $i$-th row is the feature vector $\mathbf{X}_{i}$ with length $\eta$ for node $v_{i}$, $i=1,\ldots, |\mathcal{V}|$. We assign probability values to each individual node of a graph as
\begin{equation}\label{prob-1}
     p(v_{i}) = \frac{f(v_{i})}{\sum\limits_{j=1}^{|\mathcal{V}|}f(v_{j})},
\end{equation}
where $f(v_{i})$ equals the sum of inner products between feature vector $\mathbf{X}_{i}$ and its first order neighbors' features, i.e.,
\begin{equation}\label{prob-2}
     f(v_{i}) = \sum\limits_{(v_{i}, v_{k})\in \mathcal{E}}\langle\mathbf{X}_{i}, \mathbf{X}_{{k}}\rangle.
\end{equation}
We apply the sum of feature distance between a node and its neighbors as a similarity measurement to express local features distribution. Neighboring nodes with larger inner products indicate more similarity in feature space and exhibit higher smoothness.

Relying on the definition of $p(v_{i})$ for each node, we yield the smoothness index of feature information diffusion on a graph as
\begin{equation}\label{my_graph_entropy}
     I(G) = -\sum\limits_{i = 1}^{|\mathcal{V}|} p(v_{i}) \log  p(v_{i}).
\end{equation}
It quantifies the randomness of features distribution by the ensemble average of $- \log  p(v_{i})$ over each node $v_{i}$, where $p(v_{i})$ represents the contribution of local features to the global scope in the form of probability. We could infer that features tend to scatter evenly around a graph $G$ if $ I(G)$ reaches a relatively high value.

\begin{figure*}[htbp]
\centering
\includegraphics[height=4.0cm,width=15cm]{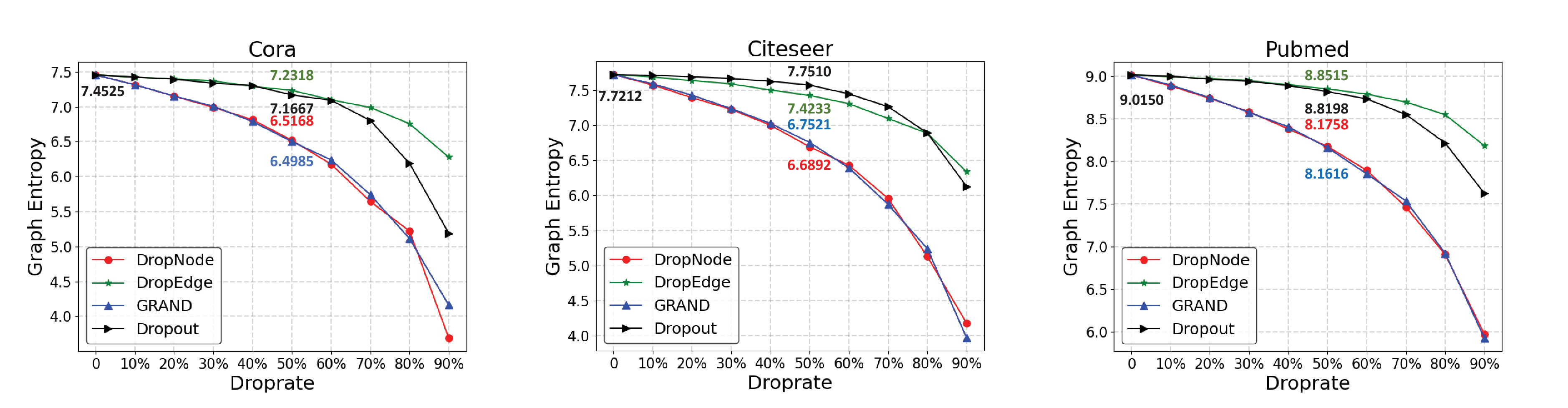}
\caption{The graph entropy curves are plotted against droprate varying from $0\%$ to $90\%$ on Cora, Citeseer, and Pubmed datasets. Each dot on curves denotes an average value over 10 times calculations.
}\label{Figure-EntropyDecaying}
\end{figure*}

In Figure~\ref{Figure-EntropyDecaying}, we take Cora, Citeseer, and Pubmed datasets as examples to show how the graph entropy varies after graph topology or features are damaged. Every curve achieves the highest graph entropy equalling 7.6357, 7.9247, and 9.6724 respectively for these three datasets. After that, these curves appear to show different performances in response to the droprate. GRAND leads to the most severe decaying on graph entropy and DropEdge gives rise to the slightest loss on graph entropy as the droprate increases. All curves decrease quickly after they meet $50\%$ droprate and stop at the lowest values at $90\%$ droprate. From the results, it is clear that all these four methods are strongly sensitive to the droprate, which reflects the damage extent of features on a graph.

\subsection{Motif-Based Information Structure}
\begin{figure}[htbp]
\centering
\includegraphics[height=2.5cm,width=8.5cm]{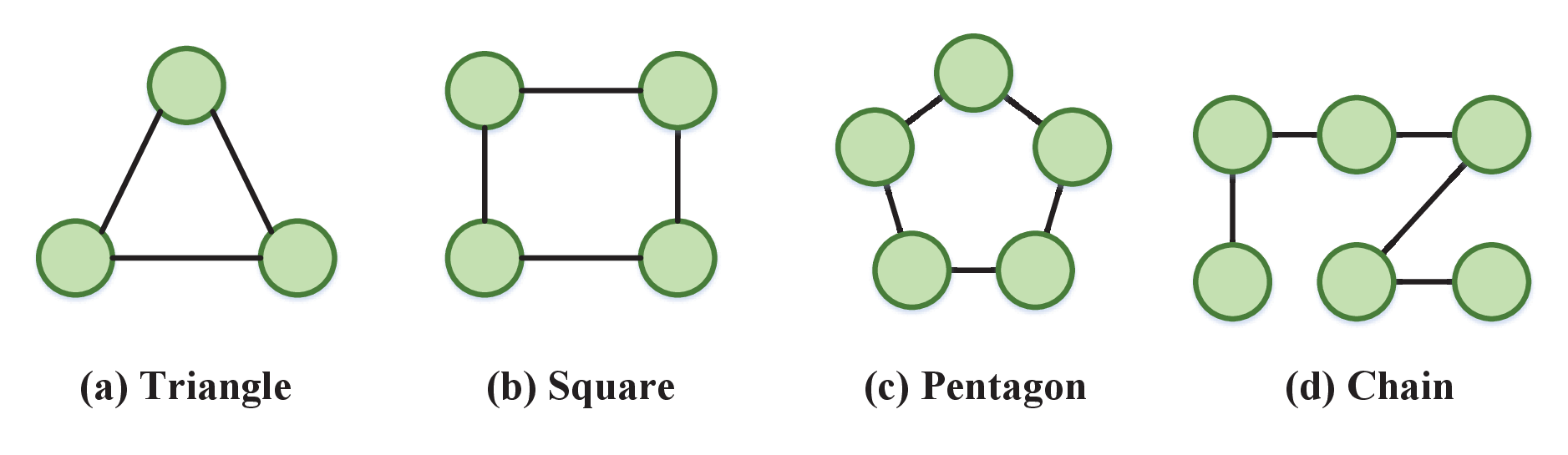}
\caption{
Examples of four typical motifs: (a) triangle motif $M_{3}^{3}$, (b) square motif $M_{4}^{4}$, (c) pentagon motif $M_{5}^{5}$, and (d) 5-hop chain motif $M_{6}^{5}$.
}
\label{Figure-motif}
\end{figure}

Next, we explore the basic multi-order information units on the data manifold. Motif-based approaches are well used in graph learning tasks, for example, community detection~\cite{motifs-EdMOT} and link prediction~\cite{motif_link_prediction}. Formally, a motif with $s$ nodes and $t$ edges can be denoted as
\begin{equation}\label{motif-Mst}
    M_{s}^{t} = ( \mathcal{V}_{M_{s}^{t}},\mathcal{E}_{M_{s}^{t}} ),
\end{equation}
where $\mathcal{V}_{M_{s}^{t}}\subseteq \mathcal{V}$ represents the set of $s$ nodes and $\mathcal{E}_{M_{s}^{t}}\subseteq \mathcal{E}$ represents the set of $t$ edges.
Several typical motifs are provided in Figure~\ref{Figure-motif}, and in particular, we focus on triangle motif $M_{3}^{3}$ in this paper.

Motifs as higher-order connectivity patterns are crucial to the construction of graph topology and the control of network behaviors~\cite{Higher-order_clustering}. In motifs, features from local dense nodes are clustered into an entirety to express information. And we define this special information structure as motif-based information structure, which shows a significant role in the preservation of graph entropy~\cite{GraphEntropyBook}.

\subsection{The Graph Entropy Preservation of Motifs}

Compared with graphs augmented by various perturbation operations, the original graph exhibits the highest entropy, since there exists no damage to both topology and features. And more precisely, the integrity of the above motif-based information structures could be completely preserved. Perturbation methods such as DropNode and DropEdge break the topological structures of motifs so that features on motifs are removed at the same time. Methods such as Dropout and GRAND do damage to the features attached to motifs without perturbing graph topology.

But different motifs show quite different effects on the preservation of graph entropy. We explore the control of entropy for triangles, squares, pentagons, and chains on Cora, Citeseer, and Pubmed graph datasets. For each dataset, we reset the feature vector on each node as zero vector except for the nodes covered in motifs, and the statistics of motifs and the derived graph entropy are reported in Table~\ref{Table-Motif-Entropy}. Compared with squares, pentagons, or chains, triangles show the advantage in the preservation of graph entropy as triangles preserve the vast majority of the original entropy, achieving 7.4016, 7.4943, and 8.9891, which are only a little bit worse than the original graph but much higher than other motif scenarios.

Triangle as a complete subgraph in graph theory or a clique in clustering algorithms, exhibits better connectivity and plays a role in the building blocks of graph topological structure~\cite{motifs-2, motifs-Graphlet}. We recall that the triangle motif acts as the basic unit for $s$-node complete subgraphs, where $s \geq 3$. Thus it is quite enough to pin the triangles to control the whole graph without the need of understanding all other higher-order complete subgraphs. With concerns about the preservation of graph entropy and the construction of graph data, we apply triangles as the basic motifs in this paper. Besides, keeping the integrity of triangle motif-based information structures could be regarded as a criterion for designing a new augmentation strategy that demands the preservation of both information and entropy.

\begin{table}
\centering
\caption{
Graph entropy calculation results derived from $5$ scenarios, which are shown in each column from left to right: the original graph, graph with only triangle, square, pentagon, or 5-hop chain motif-based information structures preserved. Here the 5-hop chains are sampled with $5\%$ ratio from the set consisting of 5-hop chains, which are derived by random walks from each source node of the graph.}
\begin{tabular}{llllll}
\toprule
Datasets    & Original  & Triangle              & Square                & Pentagon              & 5-Hop Chain   \\
\midrule
Cora        & 7.4525    & 7.4016                & 7.0189                & 7.0400                & 6.7788        \\
Citeseer    & 7.7212    & 7.4943                & 6.7188                & 6.5592                & 7.0282        \\
Pubmed      & 9.0150    & 8.9891                & 8.4884                & 8.4661                & 8.4554        \\
\bottomrule
\end{tabular}
\label{Table-Motif-Entropy}
\end{table}

\section{Methodology}\label{section-4}
\begin{figure*}[htbp]
\centering
\includegraphics[height=4.5cm,width=15cm]{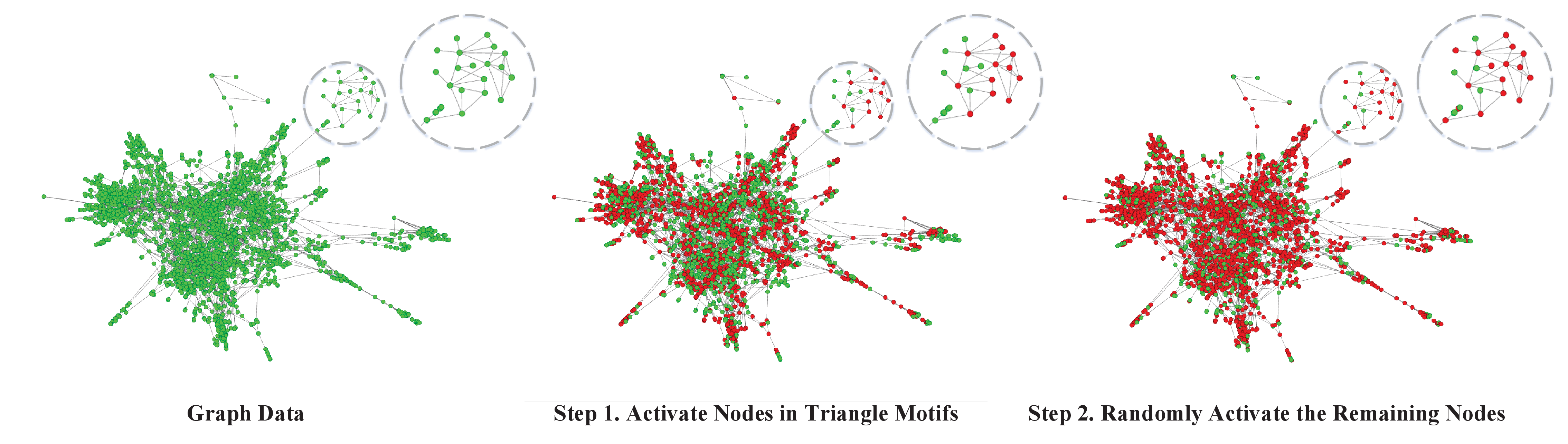}
\caption{Here we take triangle motifs $M_{3}^{3}$ to introduce our entropy preservation strategy. There are 2708 nodes and 5429 edges in total on the Cora dataset. The color of each node will turn red from green if being activated, and only features from activated nodes are retained for training. In step 1, nodes will be activated depending on whether they belong to triangle motifs $M_{3}^{3}$ or not, and the activated nodes are marked in red while others are in green. In step 2, nodes from the remaining part will be activated by Bernoulli distribution $\mathbf{B}(1-\delta)$, where $\delta = 0.5$, and then turn into red from green.
}
\label{Figure-Illustration-Cora}
\end{figure*}

Building on the above, as illustrated in Figure~\ref{Figure-Illustration-Cora}, we introduce a new graph data augmentation method with entropy preservation strategy (EP) for semi-supervised learning on graphs.

\subsection{Generate Graph Data Augmentations Using Entropy Preserving Strategy}
For a graph $G = (\mathcal{V}, \mathcal{E})$ with its adjacent matrix $\mathbf{A}$ and feature matrix $\mathbf{X}$, our method keeps the topological structure of $G$ unchanged and then takes two steps to generate multiple graph data augmentations: (1) activating nodes on motifs, (2) activating the remaining nodes with a certain probability.

In the first step, in each training epoch, we set nodes on motifs $M_{s}^{t}$ as activated status and the rest nodes as dormant status. Afterward, it generates a feature matrix $\mathbf{X}_{M_{s}^{t}}$, where only the features on activated nodes could be revealed, while the others are set as zero vectors. For the second step, we randomly sample a binary mask $\alpha_{i}$ by Bernoulli distribution $\mathbf{B}(1-\delta)$ for each node $v_{i}$ in the remaining part to determine whether $v_{i}$ would furtherly be activated or not.

To guarantee the perturbed feature vector is in expectation equal to the original vector, we multiply a coefficient $\frac{1}{1-\delta}$ and get the following as regularized perturbed feature vector
\begin{equation}\label{1}
      \mathbf{\tilde{X}}_{i} = \frac{\alpha_{i}}{1-\delta}\mathbf{X}_{i}.
\end{equation}

In summary, our proposed method generates perturbed feature matrix $ \mathbf{\tilde{X}}$ such that
\begin{equation}\label{perturbed-feature-matrix-summarized}
     \mathbf{\tilde{X}}_{i}=\left\{
     \begin{aligned}
      \mathbf{X}_{i}                                            &,  \ if \ v_{i} \in M_{s}^{t} \\
      \frac{\alpha_{i}}{1-\delta}\mathbf{X}_{i}                 &,  \ otherwise
      \end{aligned},
      \right.
\end{equation}
where $\mathbf{X}_{i}$ is the $i$-th row vector of original feature matrix $\mathbf{X}$, and binary mask $\alpha_{i}$ draws from $\mathbf{B}(1-\delta)$. The pseudo-code is shown in Algorithm \ref{Algorithm-Strategy}.

\begin{algorithm}
	\renewcommand{\algorithmicrequire}{\textbf{Input:}}
	\renewcommand{\algorithmicensure}{\textbf{Output:}}
	\caption{Graph Data Augmentation Strategy with Entropy Preservation}
	\label{Algorithm-Strategy}
	\begin{algorithmic}[1]
		\REQUIRE graph $G = (\mathcal{V}, \mathcal{E})$ and feature matrix $\mathbf{X}$, target motif $M_{s}^{t}$, Bernoulli distribution $\mathbf{B}(1-\delta)$.
		\ENSURE the augmentation of graph feature matrix: $\mathbf{\tilde{X}}$.

        \FOR {$i=1$; $i<|V|$; $i++$}
            \IF {node $v_{i}$ in $M_{s}^{t}$}
                \STATE $\mathbf{\tilde{X}}_{i} = \mathbf{X}_{i}$
                \ELSE
                    \STATE $\alpha_{i} \sim \mathbf{B}(1-\delta)$
                    \STATE $\mathbf{\tilde{X}}_{i} = \frac{\alpha_{i}}{1-\delta}\mathbf{X}_{i}$
            \ENDIF
        \ENDFOR

        \STATE \textbf{return} $\mathbf{\tilde{X}}$
	\end{algorithmic}
\end{algorithm}

In this paper, we take triangles $M_{3}^{3}$ as the basic subgraphs for motif-based information structures instead of other motifs. Supposing that the degree of each node is not greater than $m$, then the computation complexity of mining all triangles of a graph is equal to $\emph{\textrm{O}}(2 m |\mathcal{E}|)$, where $|\mathcal{E}|$ denotes the edge number. The corresponding mining procedure is provided in Algorithm~\ref{Algorithm-MineMotifs}.

\begin{algorithm}[h]
	\renewcommand{\algorithmicrequire}{\textbf{Input:}}
	\renewcommand{\algorithmicensure}{\textbf{Output:}}
	\caption{Mine Triangle Motifs}
	\label{Algorithm-MineMotifs}
	\begin{algorithmic}[1]
		\REQUIRE graph $G = (\mathcal{V}, \mathcal{E})$ where $\mathcal{V}$ denotes the node set and $\mathcal{E}$ denotes the edge set
		\ENSURE triangle motif set $\mathcal{S}$.

        \STATE $\mathcal{S} = \emptyset$
        \FOR{each edge $(v^{\prime}, v^{\prime\prime}) \in \mathcal{E}$}
            \STATE mine the neighborhood set $\mathcal{N}_{v^{\prime}}$ of node $v^{\prime}$, and $\mathcal{N}_{v^{\prime\prime}}$ of node $v^{\prime\prime}$
            \IF {$\mathcal{N}_{v^{\prime}} \bigcap \mathcal{N}_{v^{\prime\prime}} \neq \emptyset$}
                \FOR{each node $v^{\prime\prime\prime} \in \mathcal{N}_{v^{\prime}}\cap \mathcal{N}_{v^{\prime\prime}}$}
                    \STATE append $M_{3}^{3} = (\mathcal{V}_{M}, \mathcal{E}_{M})$ to $\mathcal{S}$, where $\mathcal{V}_{M_{3}^{3}} = \{v^{\prime}, v^{\prime\prime}, v^{\prime\prime\prime}\}$ and $\mathcal{E}_{M_{3}^{3}} = \{(v^{\prime}, v^{\prime\prime}),(v^{\prime}, v^{\prime\prime\prime}),(v^{\prime\prime}, v^{\prime\prime\prime})\}$
                \ENDFOR
            \ENDIF
        \ENDFOR

        \STATE \textbf{return} triangle motif set $\mathcal{S}$
	\end{algorithmic}
\end{algorithm}

\subsection{Aggregate Mixed-Order Information}
Since for various datasets, the styles of how information from multiple order neighborhoods affects central nodes are different. Hence we adopt a linear combination of different adjacent matrix powers with adaptive weights to the target dataset, i.e. $\bar{\mathbf{X}}=\bar{\mathbf{A}}\tilde{\mathbf{X}}$. Here
\begin{equation}
     \bar{\mathbf{A}}=\sum\limits_{i=0}^{d}g_{i}(\theta_{0},\theta_{1},...,\theta_{d})\hat{\mathbf{A}}^{i},
\end{equation}
is the weighted average power of symmetrically normalized adjacency matrix $\hat{\mathbf{A}}$ from order $0$ to $d$. The weight $g_{i}(\theta_{0},\theta_{1},...,\theta_{d})$ is defined by softmax function as
\begin{equation}\label{f-softmax}
     g_{i}(\theta_{0},\theta_{1},...,\theta_{d})=\frac{\exp(\theta_{i})}
  {\sum\limits_{j=0}^{d}\exp(\theta_{j})}.
\end{equation}
Note that after enough iterative calculations, parameters $\theta_{0},\theta_{1},...,\theta_{d}$ will be updated and adjusted to best values until reaching final convergence state.

\subsection{Make Prediction}
Supposing that we generate $K$ augmentations as input for each training epoch and derive perturbed feature matrix set $\{ \tilde{\mathbf{X}}^{(k)}\}_{k=1}^{K}$, each one from that set will be fed into networks $\varphi(\cdot)$ to get prediction probabilities in the form of binary matrix $\bar{\mathbf{Z}}^{(k)}\in [0, 1]^{|\mathcal{V}| \times c}$:
\begin{equation}
     \bar{\mathbf{Z}}^{(k)}= \varphi(\bar{\mathbf{X}}^{(k)}, \Omega),
\end{equation}
where $\bar{\mathbf{X}}^{(k)} = \bar{\mathbf{A}}\tilde{\mathbf{X}}^{(k)}$ and $\Omega$ denotes the parameters.

\subsection{Loss}
In the semi-supervised setting, we suppose there are $l$ labeled nodes in set $\mathcal{V}^{L} = \{v_{i}\}_{i=1}^{l}$ with their labels $\mathcal{Y}^{L} = \{\mathbf{Y}_{i}\}_{i=1}^{l}$, where $v_{i}$ corresponds to the target node and $\mathbf{Y}_{i}$ is the ground-true label, and there are $u$ unlabeled nodes in set $\mathcal{V}^{U} = \{v_{j}\}_{j=l + 1}^{l + u}$ with their labels $\mathcal{Y}^{U} = \{\mathbf{Y}_{j}\}_{j=l + 1}^{l + u}$ pending prediction. Our work follows the works of Weston et al.~\cite{deeplearning-loss} and  Feng et al.~\cite{GRAND} to design the loss function, which is a combination of the supervised loss on labeled nodes and the graph regularization loss on unlabeled nodes:
\begin{equation}
     \mathcal{L}= \mathcal{L}^{L} + \lambda \mathcal{L}^{U}.
\end{equation}

GCN model calculates each node $v_{i}$ from $\mathcal{V}^{L}$ and outputs $\bar{\mathbf{Z}}^{(k)}_{i}$ as the corresponded prediction, then derives $\mathcal{L}^{L}$ by the average cross-entropy loss over $K$ data augmentations:
\begin{equation}
     \mathcal{L}^{L} = -\frac{1}{K}\sum_{k=1}^{K}\sum\limits_{i=1}^{l}(\mathbf{Y}_{i})^\mathsf{T}\log \bar{\mathbf{Z}}_{i}^{(k)}.
\end{equation}

The graph regularization loss $\mathcal{L}^{U}$ guides the prediction of unlabeled node $v_{j}$ close to its expected label over $K$ augmentations by minimizing the distance between the prediction $\tilde{\mathbf{Z}}^{(k)}_{j}$ and $\bar{\mathbf{Z}}^{'}_{j}$,
\begin{equation}\label{equation-15}
     \mathcal{L}^{U}= \frac{1}{K}\sum_{k=1}^{K}\sum\limits_{j=l+1}^{l + u}\|\tilde{\mathbf{Z}}^{(k)}_{j} -  \bar{\mathbf{Z}}^{'}_{j}\|_{2}^{2},
\end{equation}
where $\bar{\mathbf{Z}}^{'}_{j}$ represents the possible distribution on the basis of the expected label for node $v_{j}$, and the expectation is defined in the form of
\begin{equation}\label{33}
  \bar{\mathbf{Z}}_{j} = \frac{1}{K}\sum_{k=1}^{K}\tilde{\mathbf{Z}}_{j}^{(k)}.
\end{equation}
Each $m$-th element of $\bar{\mathbf{Z}}^{'}_{j}$ refers to the probability of node $v_{j}$ on the $m$-th class, and is denoted as
\begin{equation}
     \bar{\mathbf{Z}}_{jm}^{'} = \frac{\bar{\mathbf{Z}}_{jm}^{\frac{1}{\kappa}}}{\sum\limits_{n=1}^{c}\bar{\mathbf{Z}}_{jn}^{\frac{1}{\kappa}}}, 1 \leq m \leq c,
\end{equation}
in which the categorical distribution is controlled by hyper-parameter $\kappa \in [0,1] $, and $\bar{\mathbf{Z}}^{'}_{j}$ will converge to a one-hot distribution as $\kappa$ getting close to 0~\cite{GRAND}.

\section{Theoretical Analysis}\label{section-5}
In this part, we provide a further discussion about the theoretical basis of our proposed entropy preservation strategy and its relationship with semi-supervised learning tasks for GCN. The main idea of our entropy preservation strategy originates from the smooth manifold assumption on graph data: for pairwise nodes $v_{i}$ and $v_{j}$, $i \neq j$, they share similar labels if they are close distributed on the geometry of the feature manifold. As for semi-supervised learning tasks, labeled samples commonly play roles as anchors to propagate labels to a large amount of unlabeled data, and this description refers to label propagation~\cite{label_propagation}.

We consider an $n \times n$ symmetric similarity matrix $\mathbf{W}$ on the edges of the graph $G$, such that
\begin{equation}\label{weight_matrix}
     w_{ij} = \left\{
     \begin{aligned}
      \langle \mathbf{X}_{i}, \mathbf{X}_{j}\rangle  &,  (v_{i}, v_{j})\in \mathcal{E} \\
      0                 &,  \ otherwise
      \end{aligned}.
      \right.
\end{equation}
Thus, nearby nodes in Euclidean space are assigned higher similarity if they have a larger inner product.

In the semi-supervised node classification tasks, we aim to learn a function that maps each node to its label, $\varphi: \mathcal{V} \rightarrow \mathcal{Y}$. We constrain $\varphi$ on labeled nodes to take values $\varphi(v_{i}) = \mathbf{Y}_{i}, v_{i} \in \mathcal{V}^{L}$.

Now we denote the quadratic energy function~\cite{Semi-Supervised-Harmonic} as
\begin{equation}\label{quadratic_energy_function}
\begin{split}
E(\varphi)& =\frac{1}{2}\sum\limits_{i=1}^{|\mathcal{V}|}\sum\limits_{j=1}^{|\mathcal{V}|} w_{ij} \| \varphi(v_{i}) - \varphi(v_{j}) \|_{2}^{2} \\
          & =\sum\limits_{i=1}^{|\mathcal{V}|}\sum\limits_{j=1}^{|\mathcal{V}|} w_{ij} \|\varphi(v_{i})\|_{2}^{2} - \sum\limits_{i=1}^{|\mathcal{V}|}\sum\limits_{j=1}^{|\mathcal{V}|} w_{ij}\langle\varphi(v_{i}), \varphi(v_{j}) \rangle \\
          & = \sum\limits_{i=1}^{|\mathcal{V}|} d_{i} \|\varphi(v_{i})\|_{2}^{2} - \sum\limits_{i=1}^{|\mathcal{V}|}\sum\limits_{j=1}^{|\mathcal{V}|} w_{ij}\langle\varphi(v_{i}), \varphi(v_{j}) \rangle.
\end{split}
\end{equation}

The minimum function $\varphi = \arg \min\limits_{\varphi|\varphi(v_{i})=\mathbf{Y}_{i},v_{i} \in \mathcal{V}^{L}} E(\varphi)$ of quadratic energy function $E(\varphi)$ is harmonic, as it satisfies $\Delta \varphi = 0$ for unlabeled nodes $\mathcal{V}^{U}$, i.e.,
\begin{equation}\label{harmonic}
\varphi(v_{j})  = \sum\limits_{i = 1}^{|\mathcal{V}|} \frac{w_{ij}}{d_{j}} \varphi(v_{i}) = \sum\limits_{(v_{i}, v_{j})\in \mathcal{E}} \frac{w_{ij}}{d_{j}} \varphi(v_{i}), v_{j}\in \mathcal{V}^{U}.
\end{equation}
Here $\Delta$ denotes the combinatorial Laplacian matrix in the form of $\Delta = \mathbf{D} - \mathbf{W}$, where $\mathbf{D} = diag(d_1, d_2, ..., d_{|\mathcal{V}|})$ is the diagonal matrix with each entry defined as
\begin{equation}
  d_{i} = \sum\limits_{j = 1}^{|\mathcal{V}|} w_{ij} = \sum\limits_{(v_{i}, v_{j})\in \mathcal{E}} w_{ij}.
\end{equation}

The equation (\ref{harmonic}) means the prediction of an unlabeled node is the average of $\varphi$ at neighboring nodes with the contribution coefficient $\frac{\omega_{ij}}{d_{j}}$ for each $\varphi(v_{i})$. It is worth noting that the coefficient $\frac{\omega_{ij}}{d_{j}}$ could be directly obtained from the data, while $\varphi$ denotes the map function pending learning. Thus the $\omega_{ij}$s and $d_{j}$ are the only items under our control to ensure the predictions. As for the original graph without perturbation, the item $d_{j}$ for node $v_{j}$ takes the highest value and reflects the highest smoothness on the local data manifold. However, $d_{j}$ inevitably suffers a reduction as $\omega_{ij}$ is affected by different perturbations on graph topology or features.

Now we review the definition of graph entropy $I(G)$ proposed in Section~\ref{section_Smoothness_Index}:
\begin{equation}
     I(G) = -\sum\limits_{i = 1}^{|\mathcal{V}|} p(v_{i}) \log  p(v_{i}),
\end{equation}
with the probability value of each node denoted as $p(v_{i}) = \frac{f(v_{i})}{\sum\limits_{j=1}^{|\mathcal{V}|}f(v_{j})}$, where $f(v_{i}) = \sum\limits_{(v_{i}, v_{k})\in \mathcal{E}}\langle\mathbf{X}_{i}, \mathbf{X}_{{k}}\rangle$. Based on the above notations, $p(v_{i})$ could be rewritten by
\begin{equation}\label{rewritten-p_vi}
  p(v_{i}) = \frac{d_{i}}{\sum\limits_{j=1}^{|\mathcal{V}|}d_{j}},
\end{equation}
which represents the relative contribution of local smoothness on node $v_{i}$ to the global manifold. Thus the graph entropy $I(G)$ could be is expanded as
\begin{equation}
    I(G) = -\sum\limits_{i = 1}^{|\mathcal{V}|} \frac{d_{i}}{\sum\limits_{j=1}^{|\mathcal{V}|}d_{j}} \log  \frac{d_{i}}{\sum\limits_{j=1}^{|\mathcal{V}|}d_{j}}.
\end{equation}
And in fact, the graph entropy $I(G)$ uses the form of Shannon's entropy to reflect the distribution of smoothness on the global data manifold. Compared with other perturbation methods, our entropy preservation strategy aims to preserve the global smoothness as much as possible, which is controlled by triangle motif-based information structures and quantified by graph entropy.

The smooth manifold assumption also explains the efficiency of GCN on graph data. In GCN, the recursive convolution
\begin{equation}
     \quad \mathbf{H}^{(l)}=\left\{
     \begin{aligned}
      \sigma(\hat{\mathbf{A}}\mathbf{H}^{(l-1)}\mathbf{W}^{(l-1)}) &,\ if\ l\in[1,...,k] \\
      \mathbf{X}&,\ if\ l=0
      \end{aligned}
      \right.,
\end{equation}
is to aggregate information for each node from its neighbors. This operation could be understood as a weighted sum of the neighbor features (the weights are associated with the edges) distributed on the data manifold. Thus the smoothness of a manifold is intimately connected to the amount of aggregated information for each convolution procedure of GCN, and matters to the effects of pattern recognition by GCN.

\section{Experiments}\label{section-6}
With the proposed model above, in this section, we evaluate the effectiveness of our proposed model on semi-supervised node classification tasks.

\subsection{Datasets}
We evaluate our model on real-world citation datasets Cora, Citeseer, and Pubmed~\cite{Cora-Citeseer-Pubmed}. Each citation network provides the relevant information of papers represented as nodes, each citing link between two documents by an edge, and the nodes' labels assigned by their categories. The introductions are as follows.

\begin{itemize}
  \item \textbf{Cora} contains 2708 machine learning papers divided into seven classes: Case-Based Learning Algorithm, Genetic Algorithm, Neural Networks, Probability-Based Algorithm, Reinforcement Learning, Rule Learning Algorithm, and Machine Learning Theory.
  \item \textbf{Citeseer} provides citation relationships among 3327 academic publications from an autonomous citation indexing system, which can be divided into six classes: Agents, Artificial Intelligence, Database, Information Retrieval System, Machine Learning, and Human-Computer Interaction.
  \item \textbf{Pubmed} has 19717 scientific publications about diabetes mellitus research from the Pubmed database. These publications are classified into three categories: Diabetes Mellitus Experimental, Diabetes Mellitus Type 1, and Diabetes Mellitus Type 2.
\end{itemize}

For each dataset, 1000 unlabeled nodes are selected as the test set for evaluating the classification performance. Apart from the test set, some nodes are selected by a preset partition rate $\beta = 90\%$ into the training set for learning. And among the training set, $5\%$ of the training nodes in Cora and Citeseer, $0.5 \%$ of the training nodes in Pubmed are assigned with labels to satisfy the semi-supervised setting. The data partition details are shown in Table~\ref{Table-Citation-Network}.

\begin{table*}
\centering
\caption{Statistics of the benchmark graph datasets. The columns are the name of dataset, the number of classes, the number of nodes, the number of features, the number of triangle motifs $M_{3}^{3}$, the number of nodes on $M_{3}^{3}$, the number of training nodes, the number of validation nodes and the number of test nodes.}
\begin{tabular}{lp{0.9cm}p{0.8cm}lllp{1.1cm}p{1.3cm}l}
\toprule
Datasets    &Classes     &Nodes     &Features  &Triangles      &Nodes on Triangles     &Training    &Validation & Test\\
\midrule
Cora        &7           &2708      &1433      &1630           &1470                   &1537        &171        & 1000\\
Citeseer    &6           &3327      &3703      &1547           &1183                   &2094        &233        & 1000\\
Pubmed      &3           &19717     &500       &12566          &4835                   &16845       &1872       & 1000\\
\bottomrule
\end{tabular}
\label{Table-Citation-Network}
\end{table*}


\subsection{Baselines}
To validate the performance of our approach, we compare it with a series of graph neural networks and graph augmentation methods. Here are the details of the learning methods used for comparison.

\begin{itemize}
  \item \textbf{Graph Neural Methods:} GCN~\cite{GCN} proposes convolutional architecture via a local first-order approximation containing both local graph structure and features of nodes. GAT~\cite{GAT} leverages self-attention layers to specify different weights to different nodes in the neighbors. MixHop~\cite{Mixhop} learns the mixing feature representation of neighbors at different orders. SGC~\cite{SGC} improves GCN by reducing excess complexity via removing nonlinearities and collapsing weight matrices between consecutive layers. Graph Markov Neural Network (GMNN)~\cite{GMNN} models the joint distribution of labels with a conditional random field and uses graph neural networks for classification learning. GraphSAGE~\cite{GraphSAGE} proposes a general inductive framework to embed the target node by sampling and aggregating features from its local neighborhood. FastGCN~\cite{Fastgcn} interprets graph convolutions as integral transforms of embedding function under probability measures, which are evaluated through Monte Carlo approximation.
  \item \textbf{Graph Augmentation Methods:} Graph augmentation methods used in the experiments include DropNode~\cite{AS-GCN}, DropEdge~\cite{Dropedge}, Dropout~\cite{Dropout}, and GRAND~\cite{GRAND}.
\end{itemize}

\subsection{Implementation}
In this paper, we use Python 3.7.9, Pytorch 1.0.2, Numpy 1.22.0rc2, and CUDA 10.0 as the computing environment and all experiments are conducted on the workstation with 2 INTEL XEON CPUs and 4 NVIDIA GeForce GTX1080Ti GPUs. For our proposed model, we adopt the Adam optimizer for training and initial the learning rate as 0.01 for Cora and Citeseer and as 0.05 for Pubmed. For fair evaluation, we take the same structure of a neural network with one hidden layer containing 32 neurons for both our proposed model and the baselines. And ensuring the re-productivity, the random seeds of all experiments are set to be the same values. At last, the number of training epochs is fixed to be 1000 for all datasets.


\subsection{Parameter Sensitivity and Setting}
\begin{figure}[h]
\centering
\includegraphics[height=9cm,width=9cm]{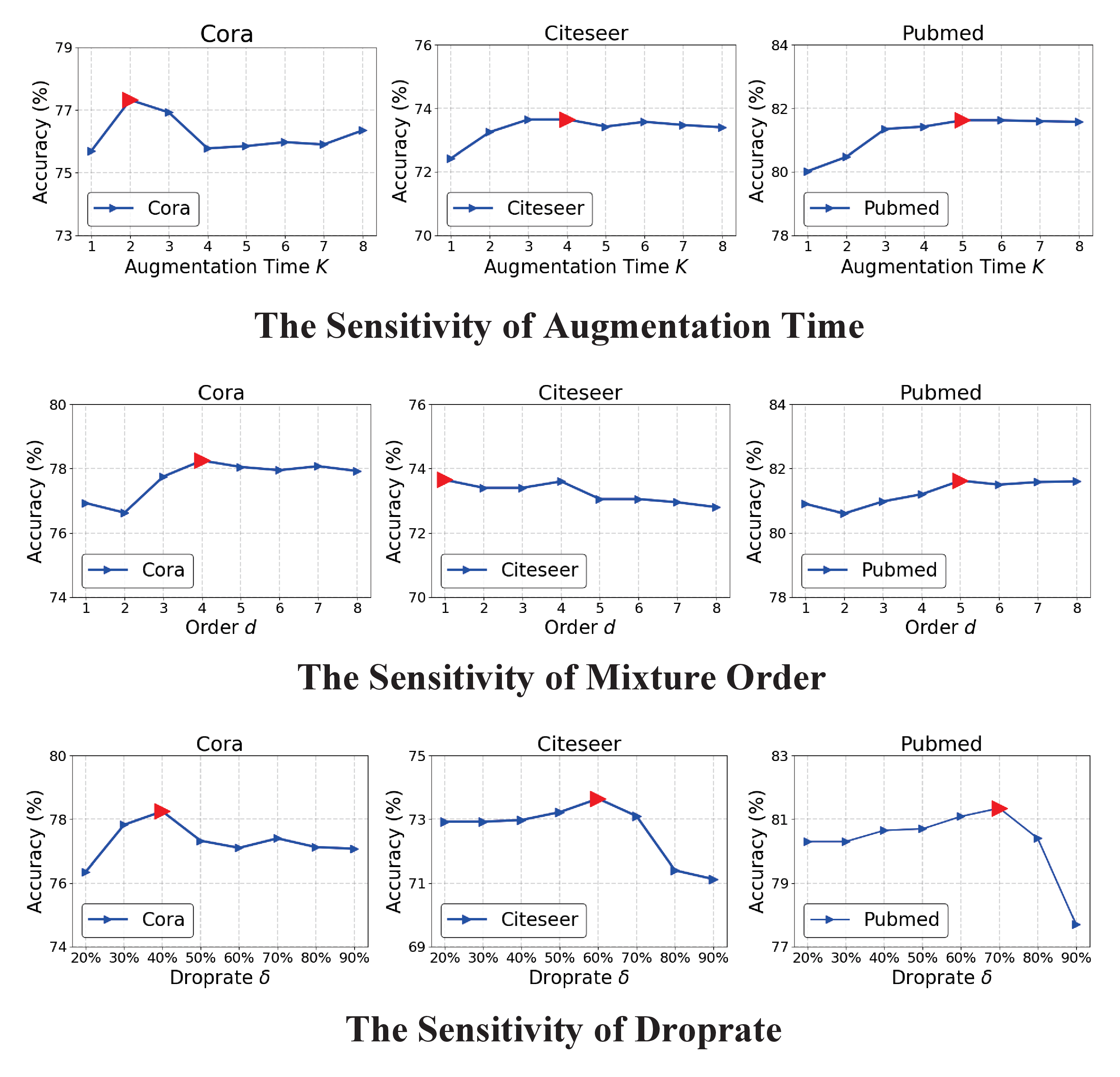}
\caption{The sensitivity of hyper-parameters.
}
\label{Figure-ParameterSensitivity}
\end{figure}

In our proposed method, there are three main hyper-parameters: augmentation times $K$, the mixture order of aggregated adjacent matrix $d$, and the droprate parameter $\delta$ in Bernoulli distribution $\mathbf{B}(1-\delta)$. In Figure~\ref{Figure-ParameterSensitivity}, we discuss the parameter sensitivity and assess how the different choices of hyper-parameters can affect our results.

We discuss the performance over different choices of augmentation times $K$ arranging from $1$ to $8$. We observe that in Cora, the performance reaches the peak when $K=2$, and then slips back to the stabilizing with accuracy equalling about $75.7\%$. As for Citeseer and Pubmed, the performances for both situations appear like a monotone trend with the increase of augmentation time $K$.

We show the classification accuracies of our model over different settings of adjacent matrix aggregated order $d$, which arranges from $1$ to $8$. In Cora and Pubmed, the performances for both situations show a monotone trend with the increase of order $d$. This implies that the aggregation of features from a larger range of the graph boosts the semi-supervised learning performance. Based on the concern of calculation efficiency, it is natural to choose the order $d$ where the highest point appears for the first time. While for Citeseer, a higher-order $d$ appears as a factor that has stunted node classification performance growth.

The parameter droprate $\delta$ in sampling distribution $\mathbf{B}(1-\delta)$ relates to the number of nodes to be activated in step 2 of our strategy. A lower droprate $\delta$ implies more nodes could be preserved as training nodes, while a larger $\delta$ brings more randomness.
In Figure~\ref{Figure-ParameterSensitivity}, it shows the trade-off between the number of training nodes and the randomness of augmentations concerning a varying $\delta$, and one can choose a better $\delta$ selectively to meet the best classification performance, i.e., $\delta=40\%$ for Cora, $\delta=60\%$ for Citeseer and $\delta=70\%$ for Pubmed.

\subsection{Comparison Results}
The comparison results of semi-supervised node classification tasks on Cora, Citeseer, and Pubmed datasets are reported in Table~\ref{Table-Results}, where the scores of our method are averaged over 10 times. We apply our entropy preservation strategy to GCN and the results of our proposed model go beyond all graph neural networks. In particular, our method gains at least $2.21\%$, $2.30\%$, and $1.16\%$ on Cora, Citeseer, and Pubmed compared with GCN. As a new data augmentation approach, our method also performs best in this category, reaching $1.55\%$, $0.25\%$, and $0.63\%$ higher in accuracy on Cora, Citeseer, and Pubmed. In addition, our method also promotes the efficiency of GAT, GraphSAGE, and FastGCN in semi-supervised classification tasks.

\begin{figure*}[htb]
\centering
\includegraphics[height=6.4cm,width=15.5cm]{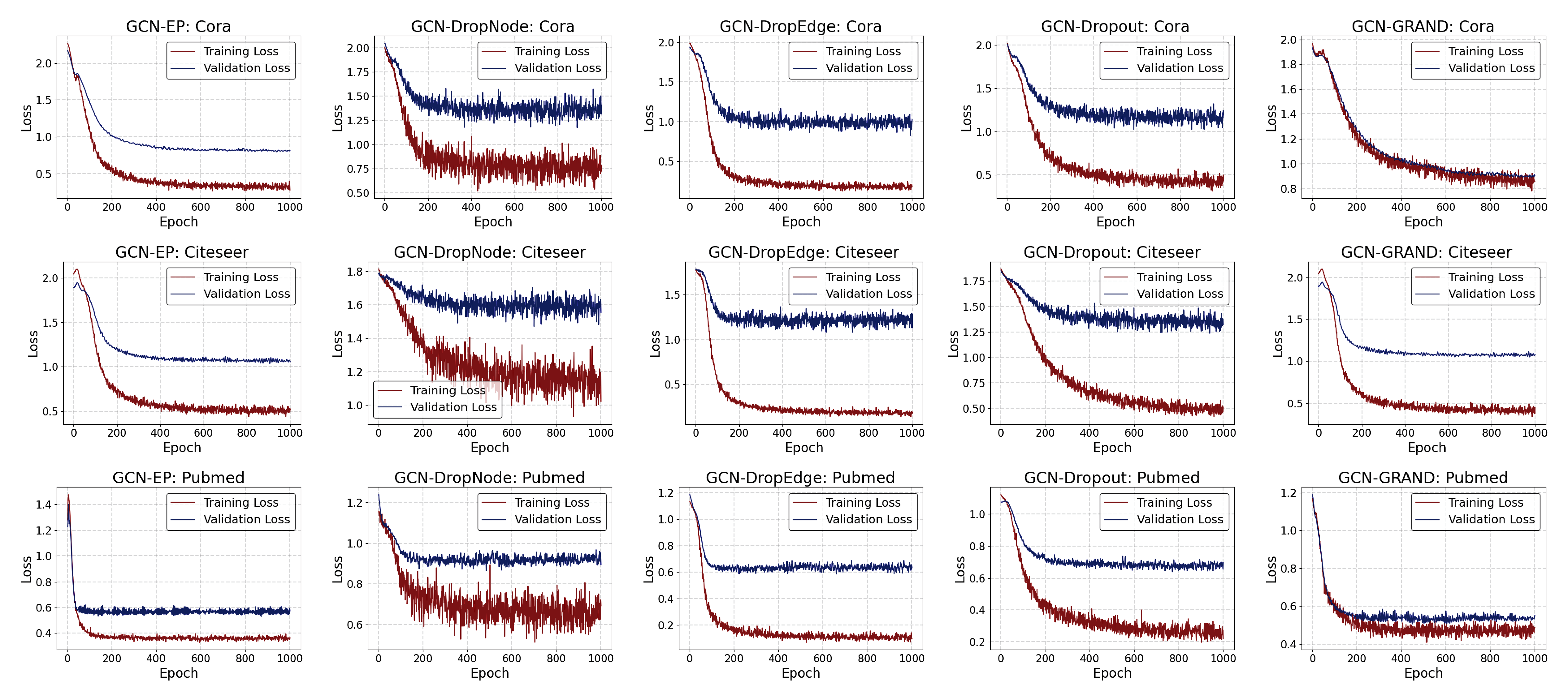}
\caption{
The loss decreasing performances of our proposed model on both training and validation set in Cora, Citeseer, and Pubmed datasets.
}
\label{Figure-5-Methods-Loss}
\end{figure*}

In Figure~\ref{Figure-5-Methods-Loss}, we utilize the decreasing performances of training and validation losses to show the training characteristic of our model compared to DropNode, DropEdge, Dropout, and GRAND. It appears that both training and validation curves of our proposed model apparently decrease smoothly and then level off at successively inferior values along with the training on all three datasets, while other methods tend to fluctuate on different levels. This obvious superiority of our model over others suggests that our model achieves more stability and robustness during the training process. Another novel finding we need to note is that for all three datasets, the decline in training or validation loss of our proposed method is faster than other methods under the same training strength and epoch number.

\begin{table}[htbp]
\centering
\caption{Semi-supervised node classiﬁcation accuracy (standard deviation) (\%) of our method and baselines on Cora, Citeseer, and Pubmed datasets.
}
\begin{tabular}{llll}
\toprule
\textbf{Algorithm}      & \textbf{Cora} & \textbf{Citeseer} & \textbf{Pubmed} \\
\midrule
GCN                     & 76.13 (0.32)      & 71.35 (0.64)      & 80.47 (0.19)  \\
GAT                     & 76.13 (0.64)      & 72.11 (0.72)      & 80.82 (0.92)  \\
MixHop                  & 77.21 (0.53)      & 71.74 (0.86)      & 80.03 (0.59)  \\
SGC                     & 76.85 (0.27)      & 71.90 (0.16)      & 79.48 (0.08)  \\
GMNN                    & 77.97 (0.46)      & 72.32 (0.88)      & 81.02 (0.34)  \\
GraphSAGE               & 76.09 (0.92)      & 68.91 (0.67)      & 78.25 (0.85)  \\
FastGCN                 & 76.80 (0.61)      & 67.33 (1.02)      & 78.26 (0.62)  \\
\midrule
GCN-DropNode            & 74.50 (2.15)  & 72.14 (3.24)      & 79.63 (1.76)  \\
GCN-DropEdge            & 76.40 (1.82)  & 71.58 (1.68)      & 80.10 (0.91)  \\
GCN-Dropout             & 76.30 (1.83)  & 71.87 (1.89)      & 80.10 (0.99)  \\
GCN-GRAND               & 76.70 (0.61)  & 73.40 (0.55)      & 81.00 (0.77)  \\
\midrule
\textbf{GCN-EP}         & 78.25 (0.42)  & 73.65 (0.38)      & 81.63 (0.54)    \\
\textbf{GAT-EP}         & 78.02 (0.61)  & 74.11 (0.69)      & 81.88 (1.14)    \\
\textbf{GraphSAGE-EP}   & 77.24 (0.78)  & 72.02 (0.91)      & 80.01 (0.62)    \\
\textbf{FastGCN-EP}     & 78.49 (0.51)  & 71.63 (0.95)      & 79.97 (0.38)    \\
\bottomrule
\end{tabular}
\label{Table-Results}
\end{table}

\subsection{Performance under Different Partition Rate}
To understand how the performance of our proposed model is sensitive to the data partition rate $\beta$, here we randomly select a fraction $\beta$ (taking value from $\{75\%, 80\%, 85\%, 90\%, 95\% \}$) of nodes that are apart from the test set into train set and $1 - \beta$ of the nodes into validation set. As shown in Table~\ref{Table-Diff-Fraction}, the node classification accuracy shows slowly monotonically increasing dependence on the partition rate $\beta$.

\begin{table}[ht]
\centering
\caption{Semi-supervised node classification accuracy ($\%$) of our proposed strategy in different partition scenarios, where partition rate $\beta$ equals $75\%, 80\%, 85\%, 90\%, 95\%$ respectively.}
\begin{tabular}{lllllll}
\toprule
Dataset         & 70\%  & 75\%  & 80\%  & 85\%  & 90\%  & 95\%  \\
\midrule
Cora            & 74.08 & 74.43 & 75.00 & 76.03 & 78.25 & 78.68 \\
Citeseer        & 71.03 & 72.17 & 72.03 & 72.23 & 73.65 & 74.10 \\
Pubmed          & 80.60 & 80.50 & 80.50 & 81.30 & 81.63 & 81.10  \\
\bottomrule
\end{tabular}
\label{Table-Diff-Fraction}
\end{table}

\subsection{Efficiency in Alleviating Over-smoothing}
\begin{figure}[ht]
\centering
\includegraphics[height=6.5cm,width=9cm]{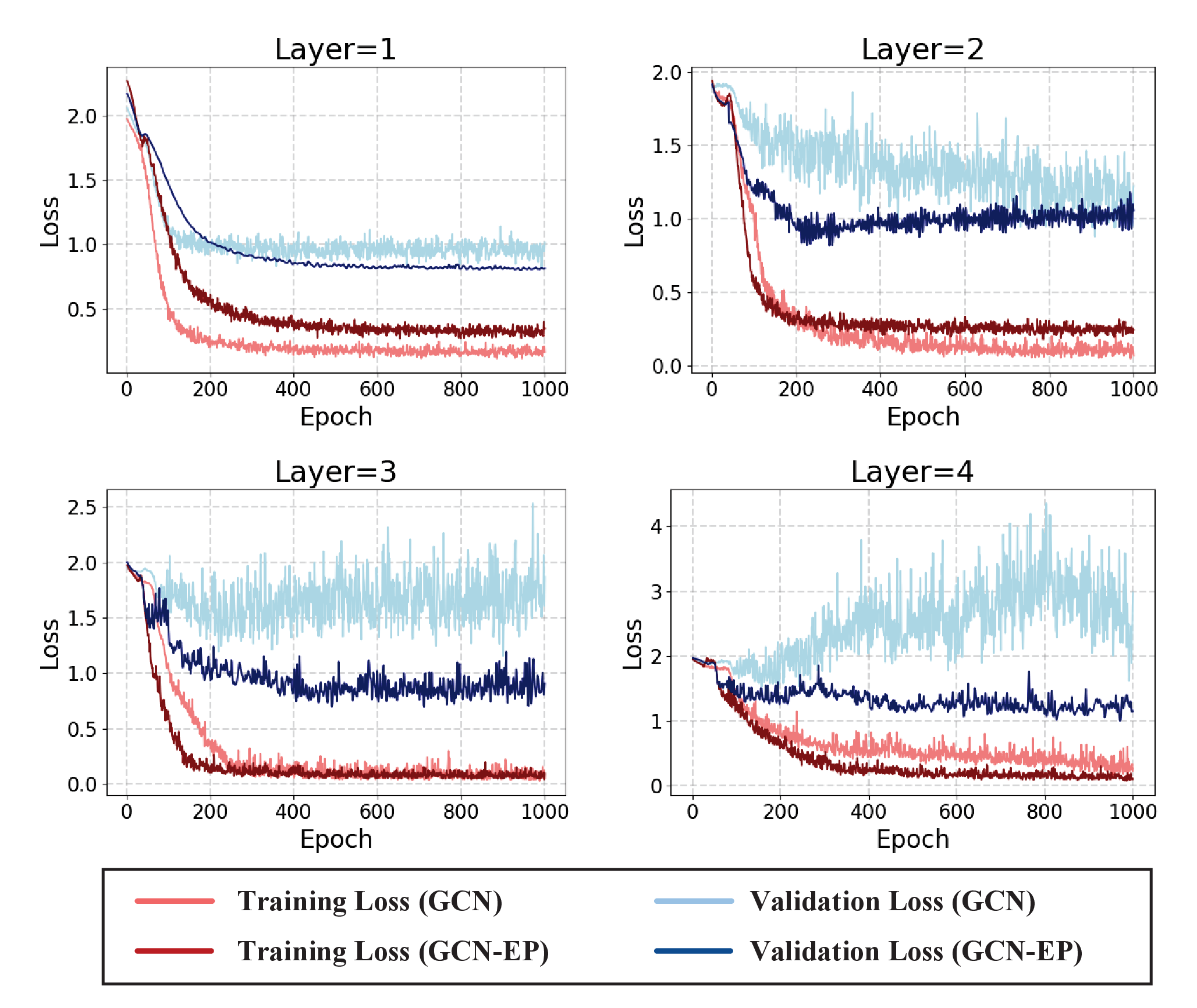}
\caption{An illustration of our entropy preserving strategy in alleviating over-smoothing phenomenon on Cora dataset.
}
\label{Figure-GCNs_over-smoothing-Loss-Alleviate}
\end{figure}

We apply entropy preservation strategy to multi-layer GCN and conduct semi-supervised node classification experiments on Cora, Citeseer, and Pubmed datasets. The performances of training and validation losses are reported in Figure~\ref{Figure-GCNs_over-smoothing-Loss-Alleviate}. As the structure of the network goes deeper, GCN-EP stabilizes both training and validation losses and greatly reduces the validation losses on all datasets. This indicts that GCN-EP could alleviate the over-smoothing phenomenon to a certain extent.

\subsection{Additional Experiments on Larger Graphs}
We also apply our proposed model on larger graphs (i.e., Facebook, Deezer, and GitHub), whose statistics are summarized in Table~\ref{Table-Larger-network}. The nodes in the Facebook dataset represent official Facebook pages while the links are mutual likes between sites. All the pages are divided into four categories: politicians, governmental organizations, television shows, and companies. In the Deezer dataset, nodes are Deezer users from European countries and edges are mutual follower relationships between them. The task related to this graph is to predict the gender of users. As for the GitHub dataset, nodes are developers who have starred at least 10 repositories and are classified based on their research directions.

\begin{table}[htbp]
\centering
\caption{Statistics of the larger graph datasets. The columns are name of the dataset, the number of classes, the number of nodes, the number of edges, the number of features, and the number of triangle motifs $M_{3}^{3}$.}
\begin{tabular}{llllll}
\toprule
Datasets    &Classes    &Nodes      &Edges      &Features  &Triangles     \\
\midrule
Facebook    &4          &22470      &171002     &4714      &797516          \\
Deezer      &2          &28281      &92752      &30978     &45034           \\
GitHub      &2          &37700      &289003     &4005      &523810          \\
\bottomrule
\end{tabular}
\label{Table-Larger-network}
\end{table}

For each dataset, we run 1000 epochs for training and then predict the labels of nodes in the test set. As shown in Table~\ref{table-results_larger_graph}, our proposed model still has gains in semi-supervised node classification accuracy compared to GCN, GAT, and MixHop.

\begin{table}[htbp]
\centering
\caption{Semi-supervised node classiﬁcation accuracy (standard deviation) (\%) of our method and baselines on larger datasets.}
\begin{tabular}{llll}
\toprule
\textbf{Algorithm}  & \textbf{Facebook}         & \textbf{Deezer}           & \textbf{Github}   \\
\midrule
GCN                 & 40.51 (0.72)              & 53.72 (0.65)              & 73.01 (1.21)      \\
GAT                 & 38.83 (0.53)              & 52.10 (0.92)              & 74.27 (2.98)      \\
MixHop              & 41.42 (0.73)              & 54.92 (0.44)              & 74.98 (1.03)      \\
\midrule
\textbf{GCN-EP}     & 42.23 (0.30)              & 56.11 (0.32)              & 75.41 (0.89)    \\
\bottomrule
\end{tabular}
\label{table-results_larger_graph}
\end{table}

\section{Conclusions}\label{section-7}
In order to tackle the phenomenon of over-smoothness and improve pattern recognition in semi-supervised learning for Graph Convolutional Networks (GCN), we propose a new graph augmentation strategy that has the advantage of entropy preservation. The theoretical basis of our proposed method lies in the smooth assumption of feature manifold, which indicates that the prediction of each targeted node is determined by its local information. To extend this smoothness from local to global and better quantify this smoothness, we introduce a new graph entropy that acts as an index to measure the distribution of global feature information. We also verify that the graph entropy is controlled by triangle motif-based information structures and note that keeping triangle motif-based information structures integrity is a very much key criterion that maintains data manifold smoothness.

Compared with other graph data augmentation methods, our strategy maintains randomness with only a small amount of graph entropy loss and without the breaking of graph topology. Several experiments on a series of graph datasets (including larger datasets) have been performed, and the results have reported improvements in terms of semi-supervised node classification tasks. A noteworthy advantage is that our method performs more stable during the whole training process, which enhances robustness. Moreover, experiments also show that our proposed method could alleviate the over-smoothing phenomenon to a certain extent.

There are a lot of interesting directions for future work. Research on graph entropy defined by different pairwise distances is still warranted for further study. In addition, using the entropy tool to investigate control problems in graph dynamics (e.g., pinning control problem~\cite{PinningControl}) is another interesting topic we aim to focus on in the future.

\subsection*{Acknowledgements}
This work is supported by the Research and Development Program of China (Grant No. 2018AAA0101100), the National Natural Science Foundation of China (Grant Nos. 62141605, 62050132), the Beijing Natural Science Foundation (Grant Nos. 1192012, Z180005).

\bibliographystyle{unsrt}
\bibliography{AGDAwithEntropyPreserving_bibfile.bib}

\end{document}